\documentclass{article}

\PassOptionsToPackage{numbers, compress}{natbib}

\usepackage[preprint]{neurips_2022}

\usepackage[utf8]{inputenc} 
\usepackage[T1]{fontenc}    
\usepackage{hyperref}       
\usepackage{url}            
\usepackage{booktabs}       
\usepackage{amsfonts}       
\usepackage{nicefrac}       
\usepackage{microtype}      
\usepackage{xcolor}         
\usepackage{bm}
\usepackage{amssymb}
\usepackage{mathtools}
\usepackage{amsmath}
\usepackage{color}
\usepackage{multirow}
\usepackage{microtype}
\usepackage{subcaption}
\usepackage{enumitem}
\usepackage{makecell, graphicx, xparse}  \usepackage[export]{adjustbox}
\usepackage{algorithm2e}
\usepackage{parskip}
\usepackage{xcolor,soul}
\usepackage{array}
\usepackage{pifont}

\SetKwInput{KwInput}{Input}                
\SetKwInput{KwOutput}{Output}  
\SetKwRepeat{Do}{do}{while}

\title{Feasible and Desirable Counterfactual Generation by Preserving Human Defined Constraints}

\author{%
  Homayun Afrabandpey \\
  Nokia Technologies\\
  Finland \\
  \texttt{homayun.afrabandpey@nokia.com} \\
  \And
  Michael Spranger \\
  Sony AI\\
  Japan \\
  \texttt{michael.spranger@sony.com}
}

\begin{document}

\maketitle

\begin{abstract}
  We present a human-in-the-loop approach to generate counterfactual (CF) explanations that preserve global and local feasibility constraints. Global feasibility constraints refer to the causal constraints that are necessary for generating \textit{actionable} CF explanation. Assuming a domain expert with knowledge on unary and binary causal constraints, our approach efficiently employs this knowledge to generate CF explanation by rejecting gradient steps that violate these constraints. Local feasibility constraints encode end-user's constraints for generating \textit{desirable} CF explanation. We extract these constraints from the end-user of the model and exploit them during CF generation via user-defined distance metric. Through user studies, we demonstrate that incorporating causal constraints during CF generation results in significantly better explanations in terms of feasibility and desirability for participants. Adopting local and global feasibility constraints simultaneously, although improves user satisfaction, does not significantly improve desirability of the participants compared to only incorporating global constraints.
\end{abstract}

\section{Introduction}

Complex Machine Learning (ML) models have been adopted in many real-world decision-making tasks either to support humans or even substitute them. Despite their superior performance, the black-box nature of these models necessitates need for interpretability methods to explain their automated decisions for individuals who are subject to these decisions. Among the large body of literature on interpretable ML \citep{burkart2021survey,tjoa2020survey}, Counterfactual (CF) explanations have shown promise for practitioners. A CF explanation contains one or more CF instances. A CF instance is a perturbed version of the original instance that flips the black--box model's prediction. By comparing a CF instance with the original instance, a human user receives hints on what changes to the current situation would have resulted in an alternative decision, i.e., ``If $\bm{X}$ was $\bm{X}'$, the outcome would have been $y'$ rather than $y$.''

Generating CF explanations that are useful in real-world is still challenging. First, CF explanations generated by many existing works do not take into account causal relationships among features. This results in CF instances that are not actionable in the real-world. Take a loan application as example, a CF explanation approach that does not adopt causal relationships among features could suggests to change the present employment type from ``newbie''\footnote{In German credit dataset, this qualitative value is defined as a person who has less than $1$ year experience in his/her current job.} to ``senior''\footnote{Between $4$ to $7$ years of experience according to German credit dataset.} while the age is unchanged. Second, CF explanations are subjective and should be personalized, while existing works do not take into account constraints from the end-users of the ML models. 
Back to the loan application example, a user might find it feasible to change the housing type, while another might instead prefer changing, e.g., number of installments (duration in month).

To bridge these gaps, we propose a new approach to generate CF explanations for any differentiable classifier via feasible perturbations. For this, we extend \citep{mothilal2019explaining} by formulating an objective function for generating CF instances that takes into account two types of feasibility constraints:
\begin{itemize}[label=\textbullet]
	\item \textbf{Global feasibilities}: unary and binary monotonic causal constraints extracted from a domain expert,
	\item \textbf{Local feasibilities}: constraints in the form of feature perturbation difficulty values, given by the end-users.
\end{itemize}
The objective function is optimized using gradient descent and feasibility constraints are satisfied during the optimization by rejecting gradient steps that do not satisfy them. It is important to note that, here we differentiate between end-user and domain expert. An end-user is the individual who is subject to the decision of the ML model, e.g. a bank customer whose loan application is rejected. A domain expert, on the other hand, knows the data and the application. We believe domain experts are naturally able to give feedback on causal relationship among (at least) some features, without being constrained to know the exact functional relationship.

The same feasibility constraints were also considered in \cite{mahajan2019preserving} for CF generation. They propose a generative model based on an encoder-decoder framework, where the encoder projects features into a latent space and the decoder generates CF instances from the latent space. Their approach, however, requires complete information about the structural causal model including the causal graph and the structural equations. This assumption is highly restrictive for applicability of the method in real-world applications. To cope with this issue, \cite{mahajan2019preserving} proposed a data driven approach to approximate unary and binary monotonic causal constraints and adopt the approximated relationships in the CF generation. For local feasibility constraints, they considered \textit{implicit} user preferences, i.e., given a pair of original instance and CF instance, $\left(\bm{x}, \bm{x}'\right)$, the user outputs $1$ if CF instance is locally feasible and $0$ otherwise. However, since there is no access to the $\left(\bm{x}, \bm{x}'\right)$ query pairs apriori, they approximate the user by first asking user preferences on some $\left(\bm{x}, \bm{q}\right)$, where $\bm{q}$ are sample CF instances generated by a CF generator without considering user preferences, and then learn a model that generates scores for each pair that mimics user preferences. 

Our approach is different from \cite{mahajan2019preserving} in several aspects:
\begin{itemize}
	\item in \cite{mahajan2019preserving}, for approximating each binary constraint, the model learns $2$ extra parameters. This hinders the scalability of the method. Furthermore, these approximated binary constraints could be imprecise as they are learned from the data, while in our approach we rely on domain experts to provide such constraints which is more reliable,
	\item local feasibility constraints are incorporated via implicit feedbacks that are approximated using a function. These feedbacks are not directly related to the final CF instances to be generated. This could result in undesirable CF instances that do not satisfy user's constraints. On the other hand, we adopt explicit user feedbacks directly into the optimization function,
	\item the type of the user feedback considered in \cite{mahajan2019preserving} for local feasibility is difficult to provide and restrictive. It is difficult to provide since the user must compare the CF instance with the original instance to find out if perturbations are locally feasible or not. It is restrictive because the approach provides no tool for the user to state the level of local infeasibilty. 
	As an example, assume a CF instance is generated by perturbing more than one feature of the original instance where all but one perturbation satisfy user's feasibility constraints. In our approach, user feedbacks are "feature level" and they are not restricted to $\{0,1\}$,
	\item last but not least, \cite{mahajan2019preserving} did not test they approach in a real user study and it is not evident from the paper how a real user could be adopted in-the-loop to obtain desirable CF explanation.
\end{itemize}

To explore the effectiveness of our explanations, we design user studies where users are asked to rank CF instances generated under different conditions. Through these studies, we found that users tend to give significantly better ranks to CF instances generated by considering global feasibility constraints compared to the case where such constraints are not considered. Furthermore, CF instances generated by adopting both local and global feasibility constraints are better than those generated by only considering global feasibility constraints. However, their difference is not statistically significant.

In summary, we make the following contributions:
\begin{itemize}[label=\textbullet]
	\item we propose a novel method to generate CF explanations that preserve both local and global feasibility constraints extracted from end-users and domain experts, respectively. This is obtained via an optimization task rather than relying on heuristics \citep{karimi2020model,tolomei2017interpretable},
	\item we design and conduct user studies to demonstrate the quality of generated CF instances. Our studies confirm that CF instances generated by capturing causal relationships are more favorable for end-users compared to those generated without causal constraints. Adding local feasibility constraints to the CF generation can further improve user satisfaction from CF instances.
\end{itemize}

\section{Counterfactual Generation With Local Feasibility via User-Defined Metrics}

Assume we want to explain the undesirable prediction $y$, of a binary black-box classifier $f$, for an instance $\bm{x}\in \mathbb{R}^{p}$. Throughout the paper, we assume that the black--box model is differentiable and does not change over time. In their seminal paper, Mothilal et. al \citep{mothilal2019explaining} proposed the following optimization function to generate a set of $K$ diverse CF instances to explain the prediction $f\left(\bm{x}\right)$:
\begin{equation}
	\arg\min_{\bm{c}_1,\ldots,\bm{c}_K} \frac{1}{K}\sum_{k=1}^{K} \mathcal{L}\left(f(\bm{c}_k),y'\right)+\frac{\lambda_1}{K}\sum_{k=1}^{K}d\left(\bm{c}_k,\bm{x}\right)-\lambda_2\mbox{d\_div}\left(\bm{c}_1,\ldots,\bm{c}_K\right),
	\label{CF_div_gen}
\end{equation}
where $\mathcal{L}$ is the loss function that pushes the predictions of $f$ for CF instances toward the desirable prediction $y'$, $d\left(.,.\right)$ is a distance measure to keep the CF instances close to $\bm{x}$, $\mbox{d\_div}\left(.,.\right)$ is the diversity metric among CF instances and $\lambda_1$ and $\lambda_2$ are regularizers that control the relative importance of diversity among the CF instances and proximity of CF instances to $\bm{x}$.

Of particular importance is the choice of distance measure $d\left(.,.\right)$ and diversity metric $\mbox{d\_div}\left(.\right)$. These metrics are subjective. When generating CF instances, a user might prefer to keep the distance between some features of the CF instance and the original instance zero, i.e., user does not tolerate any change to values of these features, while other features are adjustable. The same argument is valid for the diversity measure. To take into account these subjective constraints, we assume each user provides her preferences for perturbing features during CF generation via feature perturbation difficulty values. These values are exploited in the CF generation process via a similarity metric to be used when computing proximities. 
\subsection{Proximity}
The closeness of a set of CF instances to their original instance is equal to the mean of their negative distances.  Following \citep{wachter2017counterfactual,mothilal2019explaining}, we define separate distance metrics for continuous and categorical features.

\textbf{Continuous Features.} The distance between each continuous feature of a CF instance to the corresponding feature in the original instance can be formulated as the Mahalanobis distance with a user defined metric $\bm{A}$:
\begin{equation}
	d_{\bm{A}}^{\mbox{cont}}\left(\bm{c}_k,\bm{x}\right) = \left(\bm{c}_k-\bm{x}\right)^T\bm{A}_{\bm{\gamma}}\left(\bm{c}_k-\bm{x}\right) = \sum_{j=1}^{p_{cont}}\gamma_j\left(\bm{c}_{kj}-\bm{x}_j\right)^2
	\label{eq:mahalanobis_dist}
\end{equation}
where $\bm{A}_{\bm{\gamma}}\in \mathbb{R}^{p_{cont}\times p_{cont}}$ is a diagonal matrix with diagonal elements being encoded in the vector $\bm{\gamma}$, $\gamma_j$ is the perturbation difficulty value of the $j$\textsuperscript{th} feature, and $p_{cont}$ is the total number of continuous features. Users are restricted to assign only positive values to $\gamma_j$ to ensure positive semi-definitiy of the Mahalanobis distance metric $\bm{A}_{\bm{\gamma}}$. 
Each $\gamma_j$ indicates how much influential the feature is in determining the distance between the CF instance and the original instance; a large value for a feature implies that the feature is very difficult for the user to perturb while decreasing the value toward $1$ means that it is increasingly easier to change \citep{afrabandpey2019human}. To cancel out the effect of different feature ranges, feature-wise distances can be divided by the standard deviation of the feature. 

It has been shown in \citep{wachter2017counterfactual} that Manhattan distance normalized by the median absolute deviation of features has desirable properties. This metric can also adopt $\{\gamma_j\}_{j=1}^{p_{cont}}$ as multipliers for feature-wise distances.

\textbf{Categorical Features.} For categorical features, we use the overlap distance with feature perturbation difficulty values as:
\begin{equation}
	d_{\bm{A}}^{\mbox{cat}}\left(\bm{c}_k,\bm{x}\right)=\sum_{j}^{p_{cat}}\gamma_jI\left(\bm{c}_{kj}\neq\bm{x}_j\right),
	\label{eq:cat_dist}
\end{equation}
where $I(.)$ is the index function that returns $1$ if the condition inside it is true and 0 otherwise, and $p_{cat}$ is the total number of categorical features. Based on this metric, when $\gamma$ is large, a miss--match of the values of two categorical features results in high cost and vice versa. 

\subsection{Diversity}
Following \citep{mothilal2019explaining}, we use the determinant of the kernel matrix of CF instances:
\begin{equation}
	\mbox{d\_div}=\left|\mathbf{K}\right|,
\end{equation}
where $\mathbf{K}_{ij}=\frac{1}{1+d\left(\mathbf{c}_i,\mathbf{c}_j\right)}$, and $d\left(\mathbf{c}_i,\mathbf{c}_j\right)$ is the distance between two counterfactual instances $i$ and $j$ as defined in the previous subsection.
\subsection{Loss Function}
We use hinge loss defined as:
\begin{equation}
	\max\left(0, 1-z\times \mbox{logit}\left(f\left(\bm{c}\right)\right)\right),
	\label{eq:loss_function}
\end{equation}
where $z=-1$ for $y=0$ and $z=1$ for $y=1$ and $f(\bm{c})$ is the output of the model before entering the softmax layer (unscaled output).

\section{Modeling Global Feasibility Constraints} 
We consider the case where the structural causal model of the observed data is unknown, however there is a domain expert who is able to provide unary and binary causal constraints at least for some features. Unary constraints stipulate whether a feature can increase or decrease. For example, a unary constraint on ``Age'' restricts it to only increase since in real-world it is not possible to decrease the age. Binary constraints capture the causal relationship between two features. An example of a binary causal relationship between ``education'' and ``age'' states that increasing ``education'' results in increase of ``age''. We focus on monotonic binary constraints where increasing (decreasing) an upstream feature, causes an increase (decrease) in its child.

Given unary and binary causal constraints by the domain expert, our goal is to exploit them when generating CF instances. To model binary constraints, we define a vector $\bm{b}^i\in\mathbb{R}^p$ for each feature $i\in\{1,2,3\cdots,p\}$ that represents causal relationships where $x_i$ is the upstream variable,
\begin{equation}
	\bm{b}^i_j = \begin{cases}
		-1 & \text{if } \left(x_{i}\rightarrow x_{j}\right) \mbox{ OR } \left(i==j\right)\\
		\;0    & \text{otherwise}
	\end{cases},
\end{equation}
where $x_{i}\rightarrow x_{j}$ refers to the binary causal constraint where any change (increase/decrease) in the value of the $i$\textsuperscript{th} feature monotonically changes the value of the $j$\textsuperscript{th} feature.

Let $\mathbb{L}$ be the function to be minimized (using gradient descent) to generate CF instances. At each iteration of the optimization, 
$\nabla_{\bm{c}} \mathbb{L} = \left[\frac{\partial \mathbb{L}}{\partial \bm{c}_{1}}, \frac{\partial \mathbb{L}}{\partial \bm{c}_{2}}, \cdots, \frac{\partial \mathbb{L}}{\partial \bm{c}_{p}}\right]^T$ where the $i$\textsuperscript{th} element of the gradient vector, $\frac{\partial \mathbb{L}}{\partial \bm{c}_{i}}$, determines the direction (sign of $\frac{\partial \mathbb{L}}{\partial \bm{c}_{i}}$) and the size (magnitude of $\frac{\partial \mathbb{L}}{\partial \bm{c}_{i}}$) of the step to be taken w.r.t. the $i$\textsuperscript{th} feature to reach the minimum of $\mathbb{L}$. The combination of $\nabla_{\bm{c}} \mathbb{L}$ and $\bm{b}$ tells whether or not the gradient update violates existing binary causal constraints. This is obtained by
\begin{equation}
	\bm{v}^i = \left(\bm{b}^i\circ \mbox{sgn}\left(\nabla_{\bm{c}} \mathbb{L}\right)\right) + \left(\mbox{sgn}\left(\frac{\partial \mathbb{L}}{\partial \bm{c}_{i}}\right).\mathbf{1}\right), \;\;\; \forall i\in\{1,\cdots,p\}
	\label{eq:gradient_validity}
\end{equation}
where $\circ$ is the element-wise multiplication, $\mbox{sgn}\left(.\right)$ is the sign function, and $\mathbf{1}\in \mathbb{R}^p$ is the vector of all ones. Elements of $\bm{v}^i \in \mathbb{R}^p$ could have values in $\{0,\pm1, \pm2\}$ with the following interpretation:
\begin{itemize}[label=\textbullet]
	\item when $v^i_j=0$, its either because there is no (monotonic) causal relationship between the two features, or because the gradient signs satisfies the causal relationship, or because $i=j$,
	\item $v^i_j=\pm2$ happens when there is a causal relationship between the two features, but the gradient signs for the two features violates the causal relationship, 
	\item finally, $v^i_j=\pm1$ happens only when the gradient of the loss function w.r.t one of the features is zero. If the gradient w.r.t. the downstream feature, i.e., $x_j$, is zero, then the causal relationship is violated since the upstream feature changes while the downstream feature remains constant. Otherwise, the update is valid.
\end{itemize}

With the above interpretation and given the vectors $\{\bm{v}^i\}_{i=1}^p$, one accepts or rejects gradient update w.r.t. the $i$\textsuperscript{th} feature as follows:
\begin{multline}
	\frac{\partial \mathbb{L}}{\partial \bm{c}_{i}} = \begin{cases}
		0         & \text{if } \left(v^i_j = \pm2\right) \vee \left(v^i_j=\pm1 \wedge \frac{\partial \mathbb{L}}{\partial \bm{c}_{j}} = 0\right)\\
		\frac{\partial \mathbb{L}}{\partial \bm{c}_{i}}      & \text{otherwise}
	\end{cases}, \;\;\; \forall j\in\{1,\cdots,p\}.
	\label{eq:gradient_update}
\end{multline}

To apply unary constraints, we define two hypothetical features shown by $U^+$ and $U^-$. These can only be used as upstream features; $U^+\rightarrow x_k$ determines that $x_k$ can only increase, while $U^-\rightarrow x_k$ states that the feature can only decrease. We define vectors $\bm{b}^{+}\in\mathbb{R}^p$ and $\bm{b}^{-}\in\mathbb{R}^p$ to model unary constraints where each row of these vectors determines whether or not the feature of the corresponding row is downstream feature for $U^+$ or $U^-$, respectively. With this definition, we have $\bm{v}^{+} = \bm{b}^{+}\circ\mbox{sgn}\left(\nabla_{\bm{c}}\mathbb{L}\right)$ and $\bm{v}^{-} = \bm{b}^{-}\circ\mbox{sgn}\left(\nabla_{\bm{c}}\mathbb{L}\right)$. Both vector $\bm{v}^{+}$ and $\bm{v}^{-}$ have values in $\{0,\pm1\}$. When $\bm{v}^{+}_j = -1$, the unary constraint is violated by the update to the $j$\textsuperscript{th} feature, while $\bm{v}^{+}_j=0$ means that the update is zero and $\bm{v}^{+}_j=+1$ represents that the gradient update satisfies the constraint. For $\bm{v}^{-}$, if the value of an element equals $+1$, the constraint is violated. 

\section{Empirical Evaluation}\label{Sec:experiments}
We conduct two user studies as explained in the following subsections to verify how good are the explanations generated using our approach when compared to DiCE \cite{mothilal2019explaining}.

\subsection{Datasets and Model}\label{subsec:datasets}
The datasets used in the experiments are Adult and german credit datasets. We adopted the pre-processed version of the adult dataset based on \cite{haojun2016predicting} with $8$ features, namely, age, work class, education level, marital status, occupation, race, gender, and hours per week. For german credti, we consider $9$ demographic and socio-economic features, including: duration in month, credit history, credit amount, present employment since unemployed, sex, age, job, and number of people liable to provide maintenance for. For both datasets, categorical features are one-hot-encoded and continuous features are scaled between $0$ and $1$. Datasets are divided into $80\%-20\%$ train and test sets.

The black-box model is a single layer neural network trained for $20$ epochs with learning rate $0.01$ using ADAM optimizer \cite{kingma2015adam} to minimize cross entropy loss. Accuracies of the trained neural network on held-out test set are $83\%$ and $73\%$ for Adult and German Credit, respectively. In all experiments, we followed \cite{mothilal2019explaining} and used Manhattan distance normalized by median absolute deviation of features as distance metric for continuous features when generating counterfactual instances using DiCE and our approach. Both regularizers in Eq. \ref{CF_div_gen}, i.e., $\lambda_1$ and $\lambda_2$, are set to $1$ in the implementation. All experiments were run on a $1.9$ GHz CPU with $8$ GB RAM. 

\subsection{Baselines}
We focus on DiCE \cite{mothilal2019explaining} as baseline, because our implementation is based on the DiCE code and therefore comparison is easier. In the following subsection, for the sake of brevity, we call our proposed method C-DiCE, i.e. DiCE that also takes into account Causal constraints.

\subsection{Feasibility Constraints and User Satisfaction}\label{user_study}

Most of the existing approaches for counterfactual explanation are validated without user experiment. A key limitation of this approach is that predefined metrics such as those adopted in \cite{mahajan2019preserving,mothilal2019explaining,sharma2019certifai}, do not precisely capture the human cognition when evaluating subjective criteria such as desirability of CF instances. For this, we develop experiments with real users to study the goodness of CF explanations considering following conditions,
\begin{itemize}[label=\textbullet]
	\item \textbf{C1}: CF instances generated by DiCE, i.e., without feasibility constraints;
	\item \textbf{C2}: CF instances generated by C-DiCE by taking into account only global feasibility constraints;
	\item \textbf{C3}: CF instances generated by C-DiCE considering both global and local feasibility constraints.
\end{itemize}
Users are asked to rank CF instances generated with DiCE, C-DiCE:C2, and C-DiCE:C3, for several real instances based on three criteria: (i) validity, (ii) feasibility, and (iii) desirability. Validity determines whether or not the generated CF instance flips the outcome of the original instance. Feasibility states whether or not the values assigned to each feature or combination of features are feasible considering real-world constraints. Finally, desirability reflects users' satisfacation on CF instances. Our hypotheses were:
\begin{itemize}[label=\textbullet]
	\item \textbf{H1}: CF instances generated by C-DiCE:C2 have, on average better ranks in terms of the above-mentioned criteria, than those generated by DiCE;
	\item \textbf{H2}: C-DiCE:C3 generates CF instances with better average ranks compared to both DiCE and C-DiCE:C2 in terms of the introduced criteria.
\end{itemize}
We conduct two user studies: in the first one users were asked to rank CF instances generated by DiCE and C-DiCE:C2 using Adult dataset, and in the second study users were asked to rank CF instances generated by C-DiCE:C2 and C-DiCE:C3 using German credit dataset.

Since we do not have the true causal constraints, we asked $4$ data scientists who were familiar with the datasets to determine unary and binary causal constraints. Extracted constraints highly agreed with agreement rate above $90\%$ on average over all experts, however we only adopted constraints all experts agreed on. Constraints defined for adult dataset are: $education \rightarrow age$, $U^+ \rightarrow age$, and $U^+ \rightarrow education$. 
For German credit dataset the feasibility constraints are: $present\_employment\_since\_unemployed \rightarrow age$, $job \rightarrow age$, $U^+ \rightarrow age$.

From the test set of each dataset, we randomly assign $5$ samples with undesirable output 
to each participant. These samples are drawn randomly without replacement to ensure that participants are assigned unique samples. For each sample, we generate $10$ CF instances; in the first experiment, $5$ using DiCE and $5$ using C-DiCE:C2, and in the second experiment, $5$ using C-DiCE:C2 and $5$ using C-DiCE:C3. These CF instances are shuffled and showed to a participant together with their original instance, however, the participant is unaware of the condition using which the CF instances are generated. CF instances are visualized to the user in a spreadsheet. Developing a more user-friendly tool for visualization and feedback collection is out of the scope of this paper. The participant is asked to rank these CF instances from $1$ to $10$, where $1$ is the best rank and $10$ is the worst. Participants are not allowed to assign same ranks to several CF instances unless they are identical. At the end of the user study, we conduct an interview with the participants to discuss their ranks and ensure they understood the task correctly.
\begin{figure}[t!]
	\centering
	\begin{tabular}[b]{c}
		\includegraphics[scale=.32]{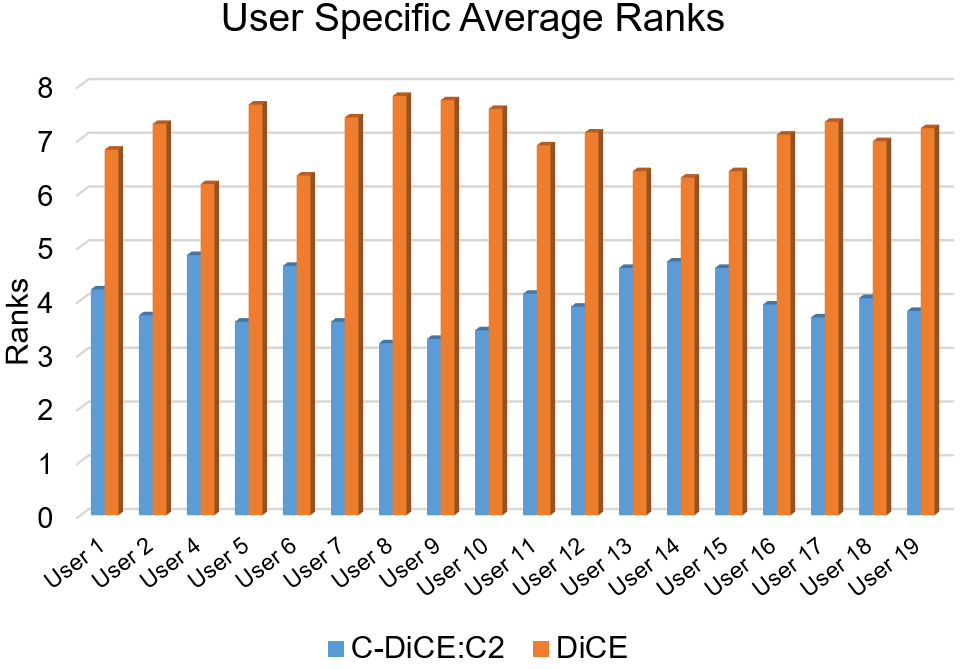} \\ (a)
	\end{tabular} \hspace{-3pt}
	\begin{tabular}[b]{c}
		\includegraphics[scale=.32]{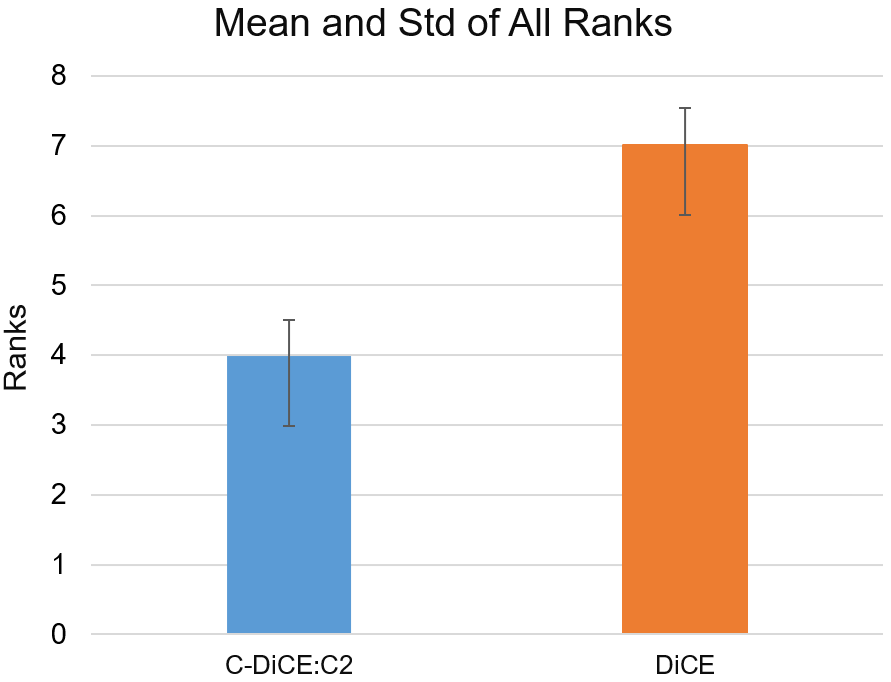} \\ (b)
	\end{tabular} \\
	\begin{tabular}[b]{c}
		\includegraphics[scale=.32]{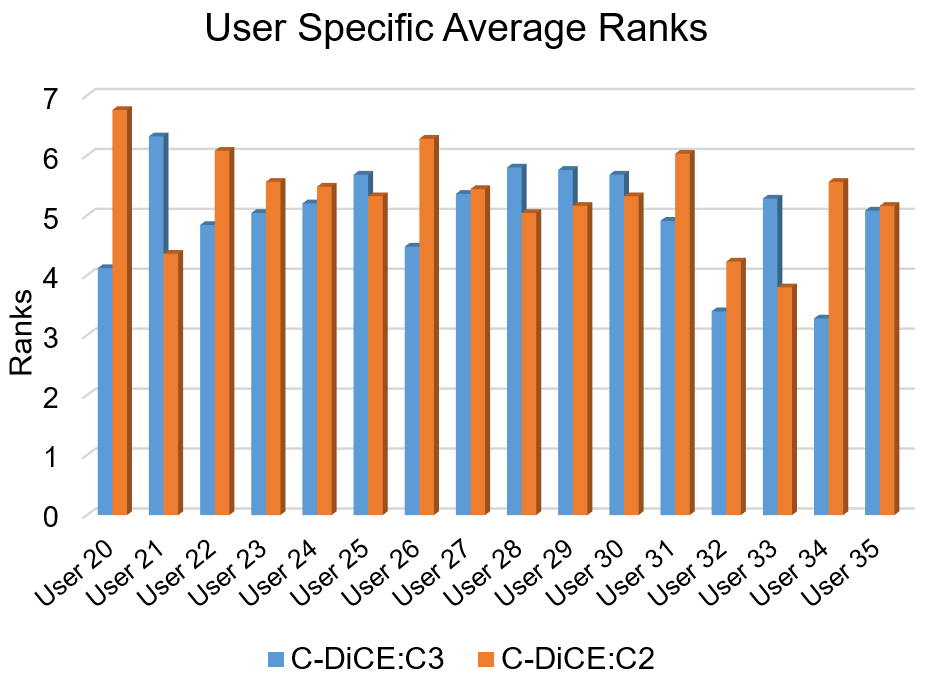} \\ (c)
	\end{tabular} \hspace{-3pt}
	\begin{tabular}[b]{c}
		\includegraphics[scale=.32]{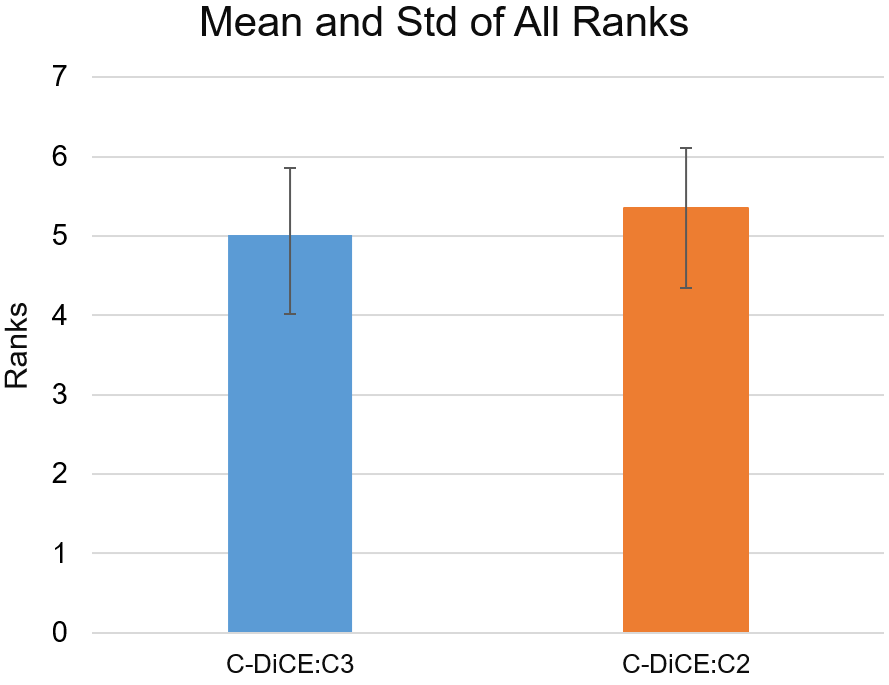} \\ (d)
	\end{tabular}
	\caption{Results of the user studies for comparing CF instances generated by (a,b) DiCE and C-DiCE:C2 and (c,d) C-DiCE:C2 and C-DiCE:C3: (a,c) mean of the ranks provided by each user, (b,d) mean and standard deviation of the ranks provided by all users. In (a) and (b) results of User 3 are discarded.}
	\label{fig:T1Results}
\end{figure}

We adopted a between subject study: $18$ participants for first experiment ($10$ male)\footnote{We hired $19$ participants for this study, but one of them (user $3$) was discarded as s/he provided incorrect input. This was pointed out during the interview.}, and $16$ participants for the second one ($9$ male). Participants were from different backgrounds including computer science ($13$), engineering fields ($8$), mathematics ($5$), social sciences ($5$), and management science ($3$). They were also from different education level including Bachelor ($6$), Masters ($8$), PhD student in 2\textsuperscript{nd} or 3\textsuperscript{rd} year ($10$), and PhD graduated ($10$); were aged $20$-$45$. Each participant was first introduced to the concept and the task using a toy example and then asked to complete the test task. To generate CF instances using C-DiCE:C3, we first ask users to provide their local feasibility constraints by assigning feature perturbation values $\gamma_j$ to each feature. In our implementation, we use $\gamma_j \in \left[1,5\right]$. Each experiment took $\sim 1$ hour and a movie ticket was awarded. 

Figure \ref{fig:T1Results}.(a,b) demonstrates results of the first user study with Adult dataset. Figure \ref{fig:T1Results}.a compares the average ranks provided by each participant to each method, DiCE and C-DiCE:C2. All users provide better ranks to CF instances generated by C-DiCE:C2. Figure \ref{fig:T1Results}.b compares mean and standard deviation of the ranks over all users. Note that average rank over all users is a value in the range $\left[3,8\right]$.

To better compare the superiority of the CF instances generated by C-DiCE:C2 compared to DiCE, Table \ref{T1:best-k} demonstrates the ratio of the top-$k$ CF instances generated by DiCE and C-DiCE:C2. According to the table, CF instances generated by C-DiCE:C2 were the top-$1$ CF instance 
almost twice as those generated by DiCE ($64.4\%$ compared to $35.6\%$). The difference is larger for top-$2$ and top-$3$ CF instances.
\begin{table}[t!]
	\centering
	\caption{Ratio (\%) of top-$k$ CF instances generated by each method to all CF instances for $k=1$, $2$, $3$ in Adult dataset. Values are averaged over $18$ users.}
	\begin{tabular}{cc|c|c}
		\cline{2-4}
		& Top-1 & Top-2 & Top-3 \\ \hline
		\multicolumn{1}{c|}{C-DiCE:C2} & $\bm{64.4}$ & $\bm{70.6}$ & $\bm{74.2}$ \\ \hline
		\multicolumn{1}{c|}{DiCE} & $35.6$ & $29.4$ & $25.8$ \\ \hline
	\end{tabular}
	\label{T1:best-k}
\end{table}

Figure \ref{fig:T1Results}.(c,d) shows the results of the second experiment. Figure \ref{fig:T1Results}.c demonstrates that $10$ out of $16$ users gave better ranks to the CF instances generated by C-DiCE:C3, where the differences between the average ranks provided by User 27 and User 35 are very marginal. Figure \ref{fig:T1Results}.d shows that average ranks over all users for the two approaches are very close to each other.

Similar to the first experiment, Table \ref{T2:best-k} demonstrate the ratio of the top-$k$ ranked CF instances generated by C-DiCE:C3 compared to those generated by C-DiCE:C2. The table demonstrates that users prefer CF instances generated by C-DiCE:C3, however the differences in the top-$k$ values are not as large as those shown in Table \ref{T1:best-k}. In this experiment, there were many identical CF instances generated by C-DiCE:C2 and C-DiCE:C3. Participants assigned same ranks to identical CF instances; consequently summation of the values of corresponding columns in Table \ref{T2:best-k} are not $1$.

\begin{figure}[t]
	\centering
	\begin{tabular}[b]{c}
		\includegraphics[scale=0.3]{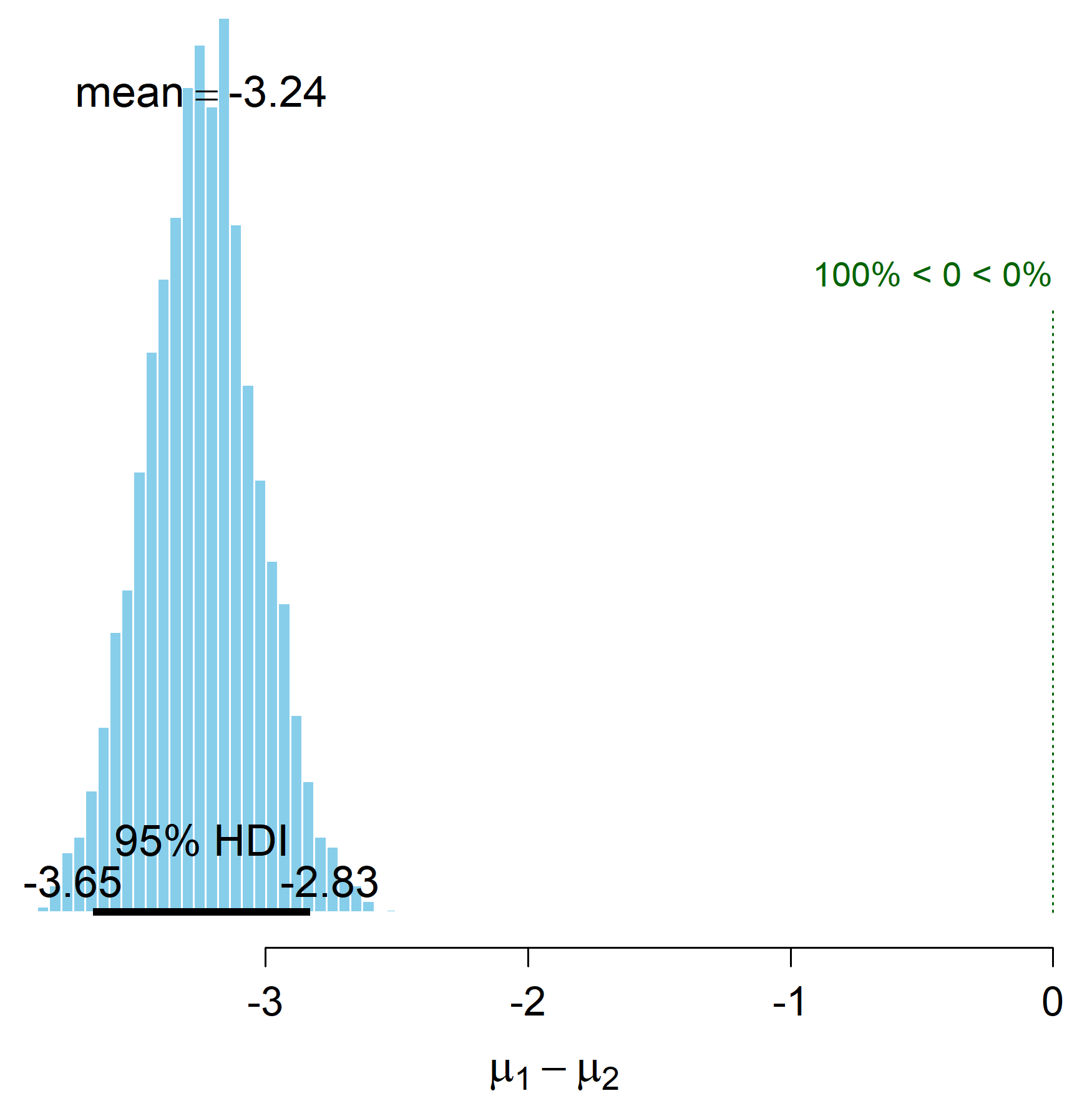} \\ (a)
	\end{tabular} \hspace{-2pt}
	\begin{tabular}[b]{c}
		\includegraphics[scale=0.3]{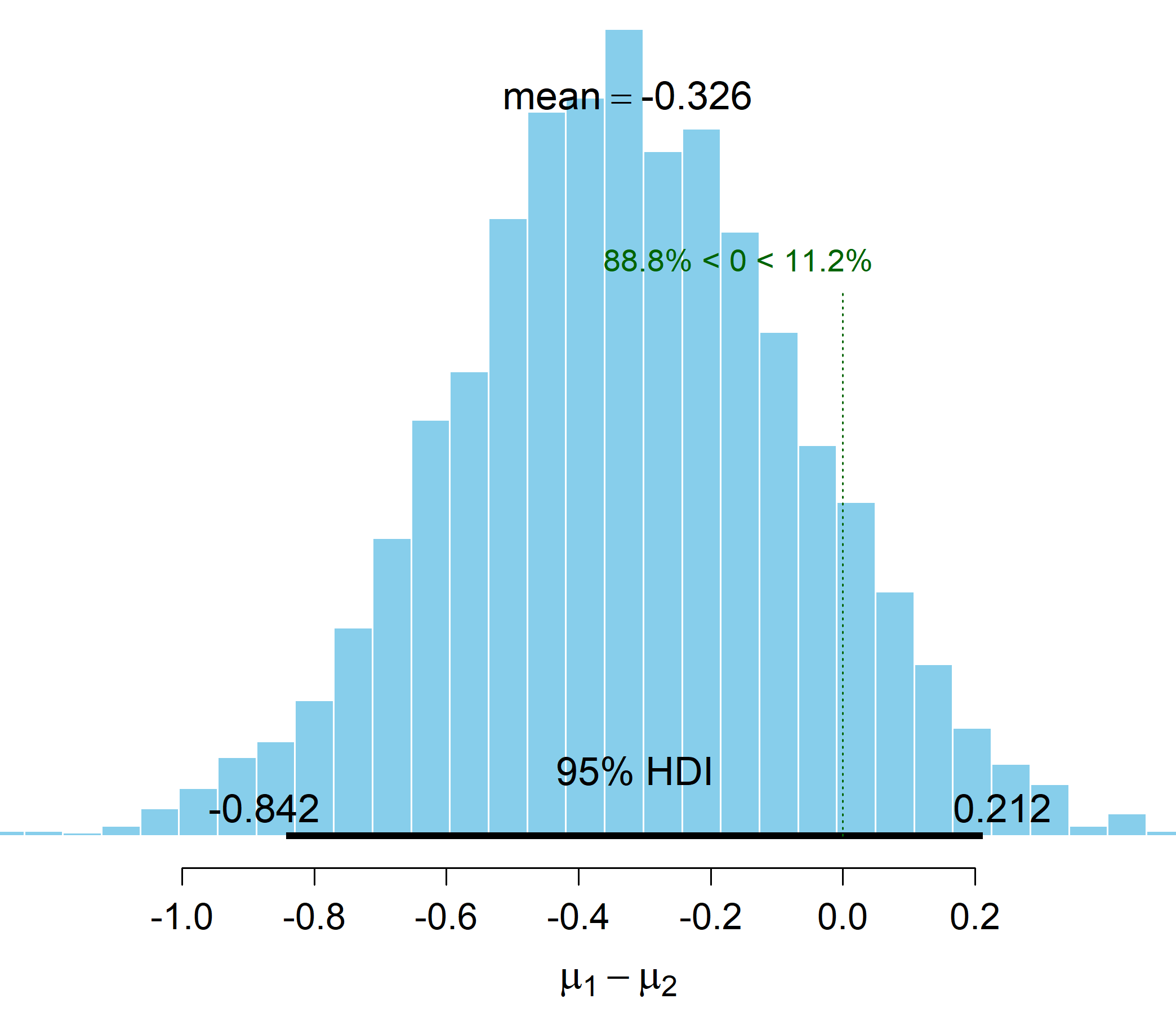} \\ (b)
	\end{tabular}
	\caption{Results of the Bayesian \textit{t}-test for the (a) first experiment where $\mu_1$ and $\mu_2$ refer to the average ranks given to the CF instances generated by C-DiCE:C2 and DiCE, respectively, and (b) second experiment where $\mu_1$ and $\mu_2$ demonstrate average ranks of the CF instances generated by C-DiCE:C3 and C-DiCE:C2, respectively. If the $95\%$ HDI contains zero, then the difference between the average ranks of the CF instances generated by the two approach is not statistically significant.}
	\label{fig:sig_test}
\end{figure}

\begin{table}[t!]
	\centering
	\caption{Ratio (\%) of top-$k$ CF instances generated by C-DiCE under two different conditions to all CF instances for $k=1$, $2$, $3$ in German credit dataset. Values are averaged over $16$ users.}
	\begin{tabular}{cc|c|c}
		\cline{2-4}
		& Top-1 & Top-2 & Top-3 \\ \hline
		\multicolumn{1}{c|}{C-DiCE:C2} & $55$ & $50.7$ & $53.5$ \\ \hline
		\multicolumn{1}{c|}{C-DiCE:C3} & $\bm{68.7}$ & $\bm{59.3}$ & $\bm{58}$ \\ \hline
	\end{tabular}
	\label{T2:best-k}
\end{table}
We use Bayesian T-test \cite{kruschke2013bayesian} to assess the significance of the differences between average ranks provided by the participants to CF instances generated by different approaches. Bayesian T-test constructs a distribution for the mean and standard deviation for the group of ranks given by the users to the CF instances generated by each approach. Then, it constructs a probability distribution over the differences between the group-specific distributions using MCMC estimation. Figure \ref{fig:sig_test} demonstrate these distributions for the (a) first experiment, where $\mu_1$ and $\mu_2$ refer to the average ranks of the CF instances generated by C-DiCE:C2 and DiCE, respectively, and (b) second experiment, where $\mu_1$ and $\mu_2$ refer to the average ranks given by the participants to CF instances generated by C-DiCE:C3 and C-DiCE:C2, respectively. Each distribution include the mean credible value as the best guess of the actual difference and the $95\%$ Highest Density Interval (HDI) as the range were the actual difference is with $95\%$ credibility. If the $95\%$ HDI includes zero, the difference between the average ranks is not significant. Figure \ref{fig:sig_test}.a, Figure \ref{fig:T1Results}, and Table \ref{T1:best-k} confirm that CF instances generated by C-DiCE:C2 are significantly better than those generated by DiCE. This difference is due to the adoption of global feasibility constraints. On the other hand, in Figure \ref{fig:sig_test}.b, the $95\%$ HDI contains zero which states that the difference between average ranks of CF instances generated by C-DiCE:C2 and C-DiCE:C3 is not statistically significant.

\section{Conclusion and Discussion}\label{conclusion}
Building upon prior work on CF explanations, we introduce a novel approach to generate feasible and desirable CF explanation. We consider two levels of feasibility, i.e., global and local. The former is defined in terms of causal constraints among variables and is extracted from domain experts, while the later captures end-user defined constraints. For global feasibility, the current work accounts for unary and binary monotonic causal constraints as two most common types of constraints and left the more complicated constraints for future work. User studies demonstrate the effectiveness of the proposed approach in increasing user satisfaction about CF explanation. Designing a tool to help domain experts to provide global feasibility constraints especially in datasets with large number of features is considered as a future work.

\bibliographystyle{abbrvnat}
\bibliography{references.bib}

\end{document}